\documentclass[]{spie}  
 \usepackage{amsmath,amsfonts,amssymb}
\usepackage{graphicx}
\usepackage{float}
\usepackage[colorlinks=true, allcolors=blue]{hyperref}

\usepackage{comment}

\title{Deep learning based automatic detection of offshore oil slicks using SAR data and contextual information}

\author[a,b]{Emna AMRI}
\author[a]{Hermann COURTEILLE}
\author[a]{Alexandre BENOIT}
\author[a]{Philippe BOLON}
\author[b]{Dominique DUBUCQ}
\author[c]{Gilles POULAIN}
\author[c]{Anthony CREDOZ}

\affil[a]{Univ. Savoie Mont Blanc, LISTIC F-74000 Annecy, France}
\affil[b]{TotalEnergies S.E. Avenue Larribau, F-64018 Pau, France}
\affil[c]{TotalEnergies S.E. 2 blvd Gobert, F-91120 Palaiseau, France}

\authorinfo{Further author information: \\E.A, H.C A.B, P.B : E-mail: emna.amri@external.totalenergies.com, (hermann.courteille, alexandre.benoit, philippe.bolon)@univ-smb.fr, Telephone: +33 (0)7 51 39 45 33. D.D, G.P, A.C : E-mail: (dominique.dubucq, gilles.poulain, anthony.credoz)@totalenergies.com, Telephone: +33 (0)6 34 49 18 34.}

\begin{document} 
\maketitle

\begin{abstract}

Ocean surface monitoring, especially oil slick detection, has become mandatory due to its importance for oil exploration and risk prevention on ecosystems. For years, the detection task has been performed manually by photo-interpreters using Synthetic Aperture Radar (SAR) images with the help of contextual data such as wind. This tedious manual work cannot handle the increasing amount of data collected by the available sensors and thus requires automation. Literature reports conventional and semi-automated detection methods that generally focus either on oil slicks originating from anthropogenic (spills) or natural (seeps) sources on limited data collections. 

As an extension, this paper presents the automation of offshore oil slicks on an extensive database with both kinds of slicks. It builds upon the slick annotations of specialized photo-interpreters on Sentinel-1 SAR data for 4 years over 3 exploration and monitoring areas worldwide. All the considered SAR images and related annotation relate to real oil slick monitoring scenarios. Further, wind estimation is systematically computed to enrich the data collection. Paper contributions are the following : (i) a performance comparison of two deep learning approaches: semantic segmentation using FC-DenseNet and instance segmentation using Mask-RCNN. (ii) the introduction of meteorological information (wind speed) is deemed valuable for oil slick detection in the performance evaluation. 

The main results of this study show the effectiveness of slick detection by deep learning approaches, in particular FC-DenseNet, which captures more than 92\% of oil instances in our test set. Furthermore, a strong correlation between model performances and contextual information such as slick size and wind speed is demonstrated in the performance evaluation. 
This work opens perspectives to design models that can fuse SAR and wind information to reduce the false alarm rate.

\end{abstract}

\keywords{Deep Learning, Image Segmentation, Neural Networks, Remote Sensing, Oil Slick Detection, SAR Images, Wind Speed.}

\section{Introduction}
\label{sec:intro}  
In the wide era of offshore data, the detection of anthropogenic or natural oil slicks has been a longstanding challenge. Their potential impact on ecosystems makes their detection a mandatory task \cite{girard2003oil}. State of the art highlights the usefulness of remote sensing data for their detection, particularly the Synthetic Aperture Radar(SAR) technology \cite{brekke2005oil}. Its main advantage is providing global coverage, independent of sunlight, weather and cloud coverage. Nevertheless, slick detection remains a challenge, particularly because of the variability in slick shape and extent. In addition, slicks can be easily confused with other natural sea surface phenomena.
Moreover, oil slicks detection depends on the wind conditions that modulate the contrast of the radar response between the oil and the surrounding sea surface. A further challenge is the ability to quickly and efficiently process the number of acquired images. Since oil slick monitoring requires high resolution and high revisit frequency, automatic detection becomes necessary to assist photo-interpreters. 

To address this, advanced models based on Deep Neural Networks (DNNs) have been proposed \cite{krestenitis2019oil}, \cite{emna2020offshore} with the prospect of higher accuracy and generalization capability compared to classical approaches \cite{angelliaume2018sar}. Different strategies can be proposed. The choice depends on the operational aims, the data, the model relevance and the post-processing strategies. This work considers an annotated dataset of real slick monitoring scenarios coupled with expert interpretation and wind estimation. We propose the following contributions: i) a comparison of two approaches for oil slick detection (slick instance detection and segmentation v.s. semantic segmentation), ii) a further performance analysis that considers contextual factors such as wind speed and slick size. To our knowledge, this is the first study evaluating the impact of these critical factors on an extensive data collection covering different world areas.

The organization of the paper is as follows: Section 2 presents the oil slicks phenomena, state of the art, and the importance of the wind information. Section 3 reviews the considered models. Section 4 describes the SAR data pre-processing and the performance metrics. A final section discusses the results and further work.

\section{Oil Slicks Detection Related Works}
\label{oilSlickPhenomena}
\subsection{Oil slicks detection from SAR data} 

\begin{figure}[htb]
\begin{minipage}[b]{1.0\linewidth}
    \centerline{
    \begin{minipage}[b]{0.3\linewidth}
    \includegraphics[width=1.3in, height=1.3in]{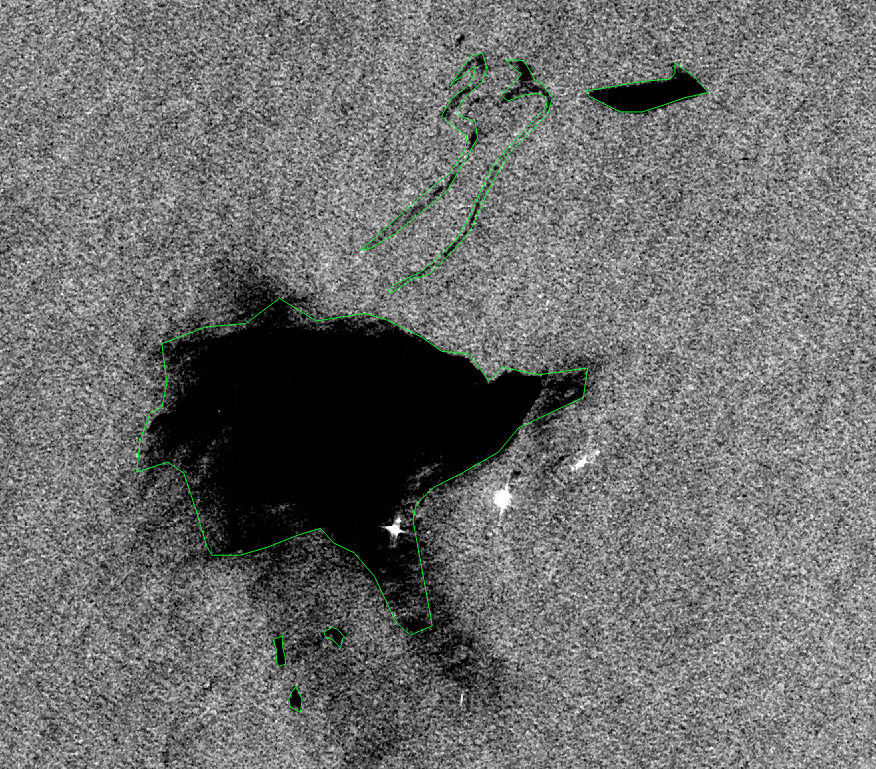}
    
    \end{minipage} 
    \begin{minipage}[b]{0.3\linewidth}
    \includegraphics[width=1.3in, height=1.3in]{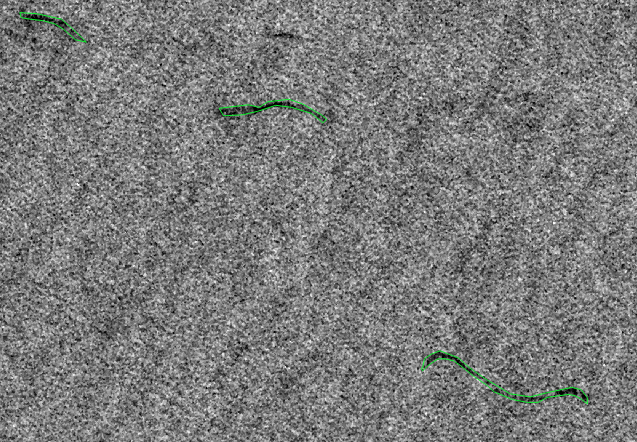}
    \end{minipage} 
    }
\end{minipage}
  \vspace{0.1in}
  \centerline{(a) Sample of oil spill (left) and seep (right) outlined in green.}
\begin{minipage}[b]{1.0\linewidth}
    \centerline{
    \begin{minipage}[b]{0.3\linewidth}
    \includegraphics[width=1.3in, height=1.3in]{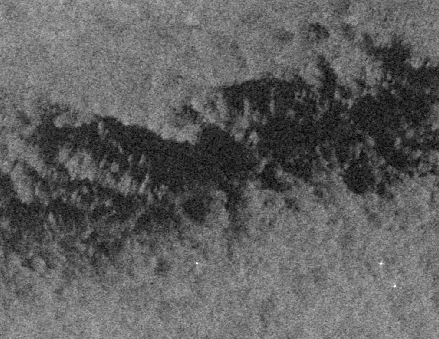}
    
    \end{minipage} 
    \begin{minipage}[b]{0.3\linewidth}
    \includegraphics[width=1.3in, height=1.3in]{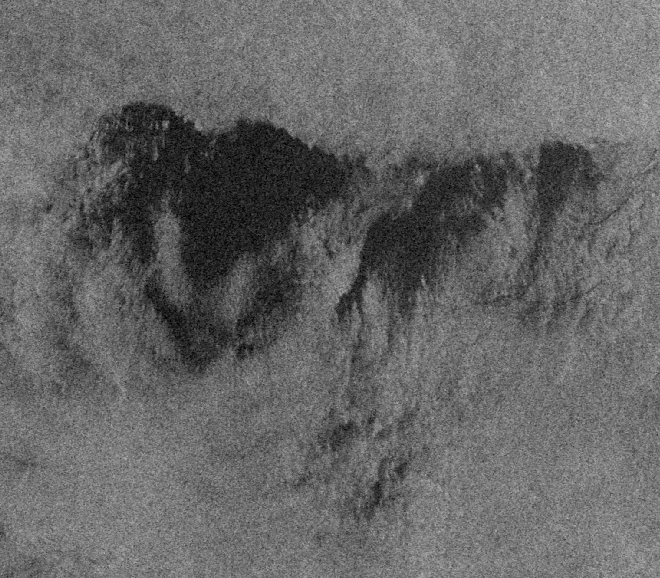}
    
    \end{minipage} 
    \begin{minipage}[b]{0.3\linewidth}   
    \includegraphics[width=1.3in, height=1.3in]{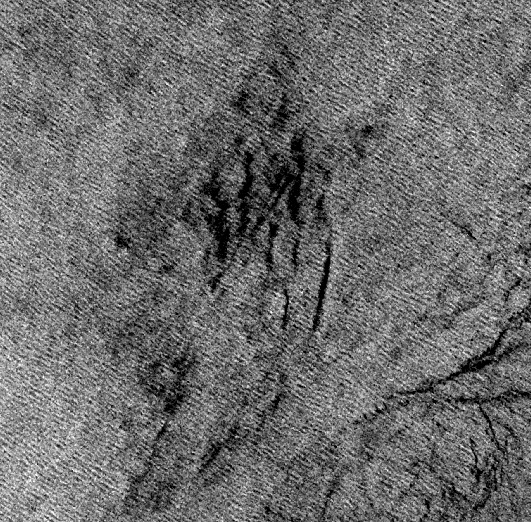}
    \end{minipage}
    }
\end{minipage}
  \centerline{(b) Lookalikes phenomenon samples, (right) low wind area, (middle) algal bloom, (left) algae.}
\vskip 0.1in
\caption{Example of oil slicks (a) and similar phenomena (b) from Sentinel-1 SAR images.}
\label{fig:sampleSliks}
\label{exemp}
\end{figure}

The phenomenon of oil slicks on the sea surface is observed relatively often. There are two types of oil slicks: anthropogenic (spills) and natural (seeps). Fig. \ref{fig:sampleSliks}(a). illustrates a sample of both types of oil slicks. Oil slicks dampen the waves on the sea surface, thus reducing the sea surface roughness and the corresponding radar back-scatter \cite{miegebielle2015use}. As a consequence, oil slicks appear on SAR images as dark patches contrasting with the brightness of the surrounding clean sea \cite{alpers2017oil}. However, several natural phenomena (algal blooms, currents and areas of low wind), called lookalikes, can generate similar areas of low back-scattering areas, as shown in Fig. \ref{fig:sampleSliks}(b). These lookalikes are therefore known to cause false alarms \cite{brekke2005oil}.

State of the art on oil slick detection is rich, we provide a brief classification of the approaches relying on the surveys proposed by Brekke et al. \cite{brekke2005oil} and Alpers et al. \cite{alpers2017oil}:\\
\begin{enumerate}
 \item Entirely manual inspection, where photo-interpreters are trained to detect and characterize oil slicks of different origins on SAR images. A tedious task with high resource costs. In addition, it cannot handle the number of images acquired by the sensors.
 
  \item Conventional approaches are based on three steps: detection and isolation of all dark patterns feature extraction \cite{liu2019semi} \cite{angelliaume2018oil} \cite{guo2017discrimination} and dark patterns classification \cite{fingas2014review}. Features may be related to wind speed, geometric parameters, contrast, proximity of ships and oil installations, and prior detection of nearby seeps in case of seeps detection. These approaches present poor generalization behaviours, remaining accurate only in specific configurations.
  
 \item  Semi-automatic approaches, where machine learning tools such as Neural Networks (NNs) are integrated into the steps of the conventional approach in order to overcome its limitations \cite{alpers2017oil}.
 
 \item End to End learning approaches such as DNNs. In this paper, particular emphasis is laid on the works carried out using SAR data.
\end{enumerate}

Different studies on offshore oil slick detection, such as Al-Ruzouq et al. \ref{al2020sensors} report several NNs models applied to oil slick detection and segmentation. Table. \ref{DLArchiWithSar} shows examples of recent NNs models using Sentinel-1 images, as well as the number and size of images used. Semantic segmentation models such as Unet and object detection methods such as Mask-RCNN have already been proposed. Moreover, it is noteworthy that few annotated SAR data are exploited, which thus limits supervised learning of large models and explains the dominance of Transfer Learning (TL) strategies that build on data representation learnt from other tasks and domains. The specificity of SAR data encourages the direct training of models with such data to avoid bias.

\begin{table}[h!]
\centering
\caption{List of different deep learning models for oil slicks segmentation using SAR data.}
\vskip 0.1in
\label{DLArchiWithSar}
  \begin{tabular}{lcccr}
  \hline
   Architecture & Crop number& Crop size& TL\\ 
  \hline
   DeepLabv3+\cite{krestenitis2019oil} & ~1002& 321×321&  -\\
   Autoencoders\cite{gallego2018segmentation}& - &  256×256&-\\
   FCN/UNet\cite{bianchi2020large} &~713 & 160×160& -\\
   Mask-RCNN\cite{emna2020offshore} & ~9302& 512×512& \checkmark \\
  \hline
  \end{tabular}
  \vskip 0.2in
\end{table}

\subsection{Contextual Data}
\label{statoftheart:wind}

Numerous studies report the interest of contextual information to better interpret SAR images for oil slick detection. The study proposed by Brekke et al. in \cite{brekke2005oil} highlights the interest in weather conditions, distance from ships, and infrastructure positions (platform, pipeline, etc.). The correlation between the visibility of oil slicks with meteorological and oceanic conditions at particular wind speeds is highlighted. They affect the slick characteristics (size, shape, etc.) while moving oil across the sea surface \cite{espedal1995oil}. Wind also impacts the back-scattering contrast between sea and slick areas. At medium wind speeds, oil wave dampening is visible but gradually disappears as wind speed increases. On the opposite, locally low wind speed, such as that created by the lee of an island, can generate lookalikes with low back-scatter \cite{wang2017sar}. During the manual annotation of the photo-interpreters, an adjustment of their assessment is made taking into account the instantaneous wind speed. According to Fingas et al. \cite{fingas2014review} guidance, a minimum wind speed of 1.5 m/s is required to allow detectability, and a maximum wind speed of 6-10 m/s will further eliminate the oil signature. The most accepted limits are 1.5 m/s to 10 m/s. Further work, such as La et al. \cite{la2018detection} and Brekke et al. \cite{brekke2005oil}, refines the wind speed range. A summary is highlighted in Table. \ref{windSpeedlevel}. However, it may remain specific to localize small-scale studies that do not permit generalization. These experiments report the requirement for moderate wind speeds.

\begin{table}
\centering
\caption{The optimal wind speed range for oil slicks detection according to the literature.}
    \label{windSpeedlevel}
    \vskip 0.1in
    \begin{tabular}{lccr}
        \hline
        Wind speed m/s & Date\\
        \hline
        3 to 7-10\cite{brekke2005oil}& 2005\\
        3.5 to 7\cite{garcia2009using}& 2009\\
        2 to 7\cite{la2018detection}& 2018\\
        \hline
    \end{tabular}
\end{table}

\section{Considered oil detection models} 

In this paper, we compare two representative neural network models among instance detection and semantic segmentation: Mask Region Based Convolutional Neural Networks (Mask-RCNN) \cite{he2017mask} with object segmentation capability and Fully Convolutional DenseNet (FC-DenseNet) for instance segmentation \cite{jegou2017one}. While the latter cannot differentiate between object instances, in the case of a two-class problem (sea/oil), a standard connected component algorithm applied on its segmentation masks adds the detection capability of the instances. Both models can thus detect oil slicks instances and related masks. However, their complexity differs strongly and impacts their training strategies, as detailed in the following. FC-DenseNet model can be trained directly on the target data. It relies on the identification of a non-linear function $f_s(x,\theta_s)=y$ with parameters $\theta_s$ optimized to map a sample SAR image $x$ to a semantic mask $y$ that differentiates the background, including lookalikes and oil slicks. It generates segmentation masks at the same image scale. This Encoder-Decoder architecture is based on dense blocks whose internal layers process the aggregation of all the preceding features. It maximizes the reuse of features and relies on significantly fewer parameters than models like the UNet vanilla model. It can thus be trained on less data with a more limited risk of over-fitting.

Regarding the Mask-RCNN model, this multitask approach generates target object region bounding boxes with object classification scores and segmentation masks. It is based on a feature extractor $f_b(x,\theta_b)=r$ that extracts the features $r$ from the $x$ input and feeds several specialized heads $f_{ti}(f_b(x,\theta_b),\theta_{ti})=y_{ti}$ so that a given head produces a task-specific output based on a unified data representation. However, its complex structure cannot be trained directly on the target data domain when few data are available. Hence the application of transfer learning is required. As stated in previous study \cite{emna2020offshore}, we transfer parameters from a pre-training on COCO dataset \cite{cocodataset} and consider the fine-tuning of the features extractor and region proposal head and the training of new classification and segmentation heads on the SAR data.

\begin{figure}[htb]
\centerline{
\begin{minipage}[b]{0.25\linewidth}
    \includegraphics[width=1.5in]{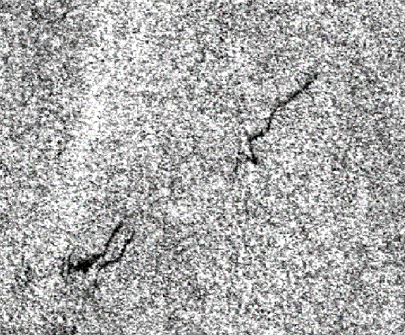}
    \leftline{(a) Input SAR image.}
\end{minipage} 
\begin{minipage}[b]{0.25\linewidth}
    \includegraphics[width=1.5in]{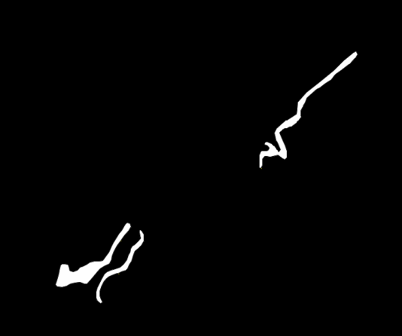}
    \centerline{(b) Output of FC-DenseNet.}
 \end{minipage} 
\begin{minipage}[b]{0.25\linewidth}   
    \centerline{\includegraphics[width=1.5in]{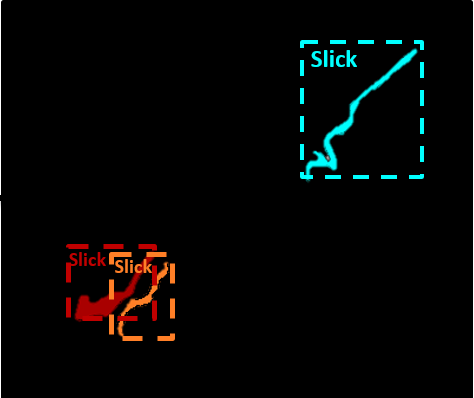}}
    \rightline{(c) Output of Mask-RCNN.}
\end{minipage}
}
\vskip 0.1in
\caption{An example of FC-DenseNet (b) and Mask-RCNN (c) prediction on a SAR image (a) containing slicks.}
\label{fig:res}
\label{exemp}
\end{figure}

\section{Used data and performance evaluation}
\label{sec:dataset}
\subsection{Data Acquisition And Pre-processing}
This work relies on ESA’s Sentinel-1 SAR data, C-band frequency (5.405 GHz) and 10m per pixel resolution. We focus on two specific areas (Namibia and Lebanon) monitored by photo-interpreters over 2015-2019. A set of 1428 images with their slick annotations is selected, and the corresponding sea surface wind estimations are provided based on the method proposed by Mouche in \cite{mouche2011sentinel}. Data pre-processing relies on the workflow described by Amri et al. in \cite{emna2020offshore} and consists of image enhancements that facilitate slicks visibility for photo-interpreters. Regarding the wind speed information, its estimation is based on empirical Geophysical Model Functions (GMFs) such as CMOD variants. It is used to extract wind speed from SAR images using the relationship between the sea surface roughness and the SAR back-scatter, considering many parameters, including incidence angle and radar wavelength \cite{mouche2011sentinel}. The wind speed is nevertheless underestimated over the slick areas because of the wave damping effect. Therefore, for performance evaluation, we consider the average wind speed on each annotated slick region from the wind speed estimated in the neighbourhood, relying on a radius of 50 meters.

The characteristics of the data crops extracted from the SAR images are summarized in Table \ref{sensortable}. The dataset is based on 512x512 pixels sized crops that are split into training and validation sets with respective ratios of 85\%/15\%, while 10 other large images are kept as a test set. Those images are captured from a third region never used for training/validation. This dataset contains 214 oil slicks and was carefully selected to give a broad representation of the slicks shape and type variation. Several lookalikes phenomena are also present, such as no wind areas. The cost of such data preparation limits the number of available samples but represents real monitoring and annotation scenarios. Since oil slicks are small targets on large images, they cover less than 4\% of the total pixels in the dataset. Thus, the imbalance between the number of slick pixels and the number of sea pixels that include lookalikes is highlighted. 

Fig. \ref{fig:resultat_wind} and Fig. \ref{fig:resultat_size} illustrate the oil slicks annotation distribution with respect to the wind speed and oil slick size ranges on the accumulated (blue and green) columns for the test. As for the training and validation dataset, small to medium sized slicks are reported. Most of them were annotated at average wind speeds (3m/s), few were detected at very low speeds and no slicks could be annotated at winds above 6m/s.


\begin{table}
\caption{Considered data collections.}
\vskip 0.1in
\begin{center}
    \begin{tabular}{l|c|c||c}
    \hline
    & Slicks& Sea including lookalikes& Total\\
    \hline
    Crops & 2208 & 861 & 3069 \\
    \hline  
    Pixels & $\sim 2*10^7$ & $\sim 7*10^8$ & $\sim 7*10^8$\\
    \hline
    Ratio & 3.4\% & 96.6\% & -\\
    \hline
    \end{tabular}
\end{center}
\label{sensortable}
\vskip -0.23in
\end{table}

\subsection{Performance Metrics}
Since both approaches allow instance-level detection and semantic segmentation, the following metrics are considered: (i) semantic segmentation quality metrics (pixel level) such as Intersection over Union (IoU) of the oil slicks, (ii) standard object detection metrics (instance level) counting detection from a minimum intersection with the Ground Truth (GT), and therefore missed detections, False Alarms (FA) and their rates. These measures are evaluated based on contextual information: slick size and local wind speed. For each, a categorization is established. For the wind, a compromise is considered between the speed ranges proposed in the literature. For slick size ranges, the categorization is made according to the size of the slick in the data collection. One must note that our study relies on real monitoring scenarios. Under these conditions, the photo-interpreter cannot provide accurate annotations on object boundaries due to fuzzy contours of the slicks and the annotation rhythm. Thus, object detection metrics are more relevant in this case study.

\section{Results and Evaluation}

These results are obtained on a test data set including several slicks of different nature, sizes and acquisition data. Besides, the test images are sourced from a new acquisition area, which is not seen during the learning phase. Changing the study area leads to variations in weather conditions and, therefore, sea surface and slicks characteristics.
The performance results of both approaches are reported with respect to the wind speed level and the size of the slick. \\

\begin{figure}[h!]
    \centering
        \includegraphics[width=0.8\linewidth]{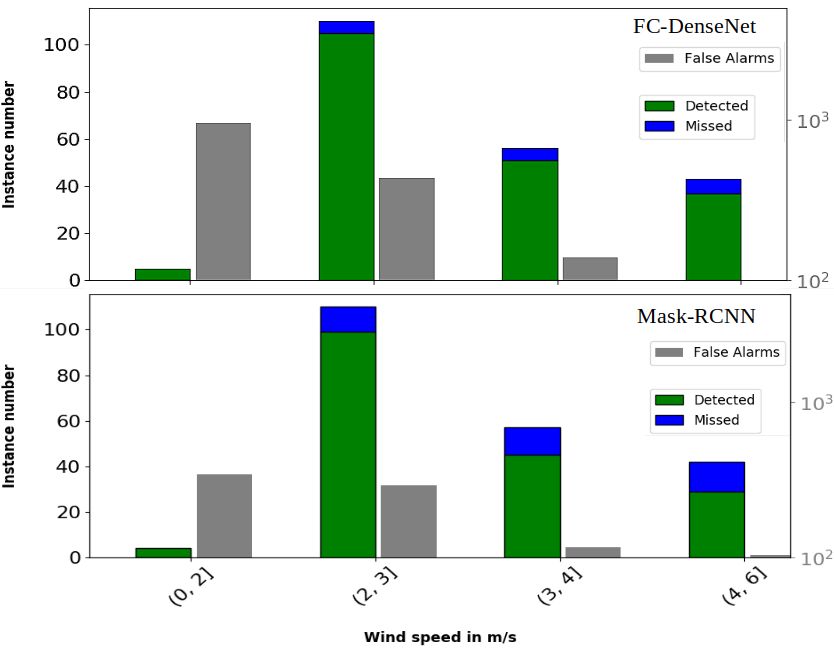}
    \caption{Slick instance detection performance with respect to the wind speed(m/s). Green for well detection, blue for missed detection and gray for false alarms (the right log-scale axis).}
    \label{fig:resultat_wind}
    \vskip -0.3in
\end{figure}

\begin{figure}[H]
    \centering
    \includegraphics[width=0.8\linewidth]{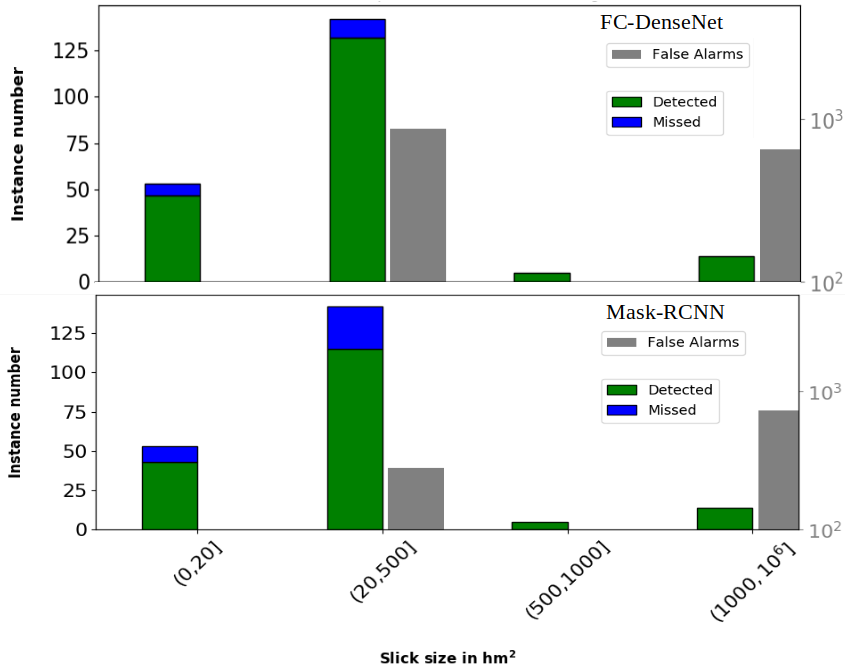}
    \caption{Slick instance detection performance with respect to the size of the GT (hm²), same legend as Fig. \ref{fig:resultat_wind}}
    \label{fig:resultat_size}
    \vskip -0.in
\end{figure}

\textbf{Quantitative results} at the oil slick instance level are presented in Fig. \ref{fig:resultat_wind} and Fig. \ref{fig:resultat_size}. Similar behaviour can be noted for both models, which are able to detect slicks over the whole range of wind speed (i.e. ]0; 6] m/s) and slick size (i.e. ]0; $10^6$] hm²). The FC-DenseNet model has a better slick instance detection rate (92.5\%) than Mask-RCNN (82.7\%). This result may highlight the difference in the training process of the models; FC-DenseNet is trained directly on the SAR data while TL of multimedia data tunes Mask-RCNN. However, the FA rate of FC-DenseNet model is 1.5 higher than that of the Mask-RCNN. For both models, FA areas are greater than 20 hm² and mostly correspond to windless areas. Furthermore, FA rates decrease as wind speed increases for both models.\\

\begin{figure}
    \begin{minipage}[b]{0.5\linewidth}
        \includegraphics[width=3.3in]{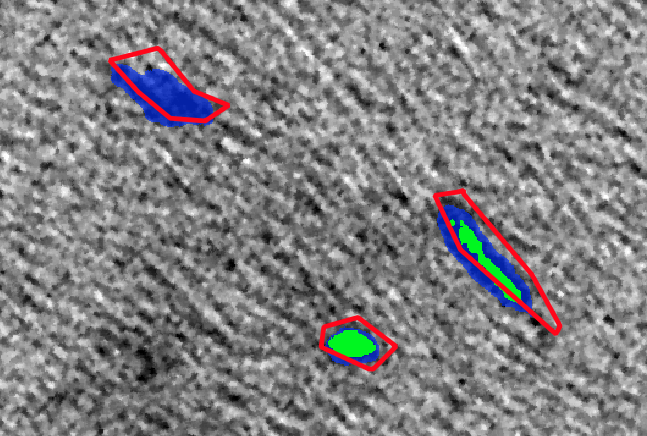}
    \end{minipage} 
    \begin{minipage}[b]{0.4\linewidth}
        \includegraphics[width=3.3in]{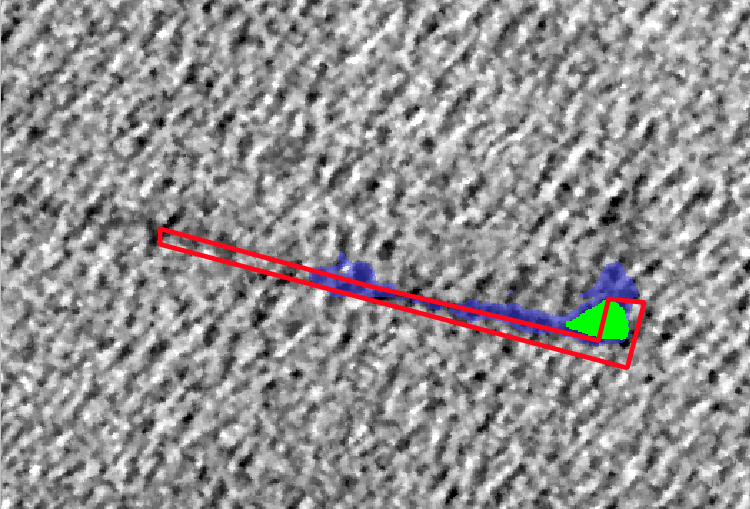}
     \end{minipage} 
    \caption{An example of FC-DenseNet (blue) and Mask-RCNN (green) prediction on Sentinel-1 images where the photo-interpretation slick annotation is marked by red polygons.}
    \label{fig:example_prediction}
    \vskip -0.1in
\end{figure}

More globally, over the entire test dataset, the amount of well detected slick pixels for FC-DenseNet (13\%) is significantly better than for Mask-RCNN (7\%) with slick IoU values of 0.30 and 0.06 respectively. The slick IoU values remain low, which can be explained partly by the annotation uncertainty of the slick boundaries during the manual annotation step that is shown in Fig. \ref{fig:example_prediction}.

\textbf{Qualitative results} are illustrated in Fig. \ref{fig:example_prediction}. Four oil slicks have been selected from the test set. Both model predictions are reported as well as the GT annotations. One observes that Mask-RCNN omits one annotated oil slick while FC-DenseNet finds them all. The IoU results can be illustrated in Fig. \ref{fig:example_prediction}, where Mask-RCNN shows less overlap with the GT once the detection is valid. Regarding FC-DenseNet, the higher FA rate can be explained visually: oil slicks masks can be fragmented into several instances that do not always intersect with the raw annotation. A post-processing step and refined annotation can solve such a problem.



We notice that in the case of large slicks, the detection of FC-DenseNet is fragmented, as shown in the right prediction of Fig. \ref{fig:example_prediction}. This may be caused by the lack of large slicks in the database, as they are divided into (512*512) crops, so the network is given a partial view. If a slick covers the entire crop, its detection will be hampered by the lack of contextual information. Morphological post-processing can be performed to optimize the detection of large slicks.


\section{Conclusion and Future works}
This paper compared instance detection and semantic segmentation approaches for oil slick detection relying on popular deep neural network models. The contribution focuses on performance analysis regarding contextual information, particularly wind speed and slick size in a real oil slick monitoring scenario. Offering readymade predictions to the photo-interpreters presents a potential change in their task. This work transforms the manual and tedious task of oil slick detection on SAR images into a task of verification and validation of the obtained predictions, thus significantly improving the process. Future work will introduce contextual information as input of the models to improve detection performance and decrease false alarm rates.

\acknowledgments 
This work has been done thanks to the HPC resources (Pangea III) from Total and GENCI-IDRIS (Grant 2020-AD011011418). We also thank NVIDIA Corporation for the donation of one NVIDIA TitanXP GPU.

\bibliography{mybib} 
\bibliographystyle{spiebib} 

\end{document}